\documentclass{Interspeech}
\usepackage{float}
 
\interspeechcameraready 

\title{ADI-20: Arabic Dialect Identification dataset and models}

\author[affiliation={1,2}]{Haroun}{Elleuch}
\author[affiliation={1}]{Salima}{Mdhaffar}
\author[affiliation={1}]{Yannick}{Estève}
\author[affiliation={1,2}]{Fethi}{Bougares}

\affiliation{LIA}{Avignon Université}{France}
\affiliation{}{Elyadata}{France}

\email{haroun.elleuch@elyadata.com, salima.mdhaffar@univ-avignon.fr, yannick.estève@univ-avignon.fr fethi.bougares@elyadata.com}
\keywords{Arabic Dialect Identification, Speech Processing, Low-resource Languages, corpus}

\usepackage{comment}
\usepackage{amssymb}
\usepackage{pifont}
\usepackage{hyperref}

\usepackage{booktabs}
\usepackage[normalem]{ulem}
\useunder{\uline}{\ul}{}
\usepackage{diagbox}
\usepackage[table,xcdraw]{xcolor}

\usepackage{pgfplots}
\usepackage{pgfplots}\pgfplotsset{compat=1.18}
\usepackage{pgfplotstable}
\usetikzlibrary{patterns}

\begin{document}

\maketitle

\begin{abstract}

We present ADI-20, an extension of the previously published ADI-17 Arabic Dialect Identification (ADI) dataset. ADI-20 covers all Arabic-speaking countries' dialects. It comprises 3,556 hours from 19 Arabic dialects in addition to Modern Standard Arabic (MSA). We used this dataset to train and evaluate various state-of-the-art ADI systems. We explored fine-tuning pre-trained ECAPA-TDNN-based models, as well as  Whisper encoder blocks coupled with an attention pooling layer and a classification dense layer. We investigated the effect of (i) training data size and (ii) the model's number of parameters on identification performance. Our results show a small decrease in F1 score while using only 30\% of the original training data.
We open-source our collected data and trained models to enable the reproduction of our work, as well as support further research in ADI.\\

\end{abstract}

\section{Introduction}
Dialect identification~\cite{pawar975comparative} is a crucial task in various natural language processing (NLP) and speech processing applications such as machine translation~\cite{salloum2014sentence}, automatic speech recognition (ASR)~\cite{waheed2023voxarabica}, sentiment analysis~\cite{kaseb2023saids}, Text-To-Speech~\cite{doan2024towards}, text normalization \cite{alhafni2024exploiting}, etc.
In everyday life, dialect is the most widely used form of communication.
Dialects are geographical and social variants of a language, shaped by syntactic, lexical, and phonological differences, including pronunciation shifts (accent).
For instance, languages such as Arabic, Chinese, English, and Spanish have numerous dialects that vary significantly across different regions and social groups, illustrating the complexity of dialect identification.
Arabic dialects, in particular, are highly diverse, with significant variations in pronunciation, vocabulary, and grammar across regions. 
Unlike European dialects, which may differ mainly in accent or vocabulary, Arabic dialects can alter fundamental aspects of the language.
Additionally, Arabic speakers frequently code-switch between local dialects, Modern Standard Arabic (MSA), and some foreign languages, making the task of dialect discrimination even more challenging.
Arabic dialects can be categorized into country-level and city-level variations \cite{ferguson2003diglossia}, reflecting both broad regional linguistic patterns and localized speech characteristics. Country-level dialects correspond to the general linguistic traits spoken across an entire nation. However, city-level dialects exist within each country, which exhibit even finer linguistic distinctions influenced by historical, social, and geographical factors.\\ 

Recently, the Arabic Dialect Identification (ADI) from speech task has gained significant attention, leading to the development of various datasets and models aimed at addressing its challenges~\cite{waheed2023voxarabica,adi5,kulkarniAnotherModelArabic2023,sullivanRobustnessArabicSpeech2023}. 
Several research initiatives have advanced ADI from speech, notably with the organization of multiple shared tasks.
The ADI shared task was introduced in VarDial 2016\footnote{Workshop on NLP for Similar Languages, Varieties, and Dialects}~\cite{malmasi2016discriminating}.
It consists of identifying a set of four regional Arabic dialects (Egyptian, Gulf, Levantine, North African) and MSA in a transcribed speech corpus. In its second edition, the task was further enriched with the inclusion of acoustic features and the release of audio files, providing richer resources for dialect classification of speech.
Driven by the success of the VarDial challenge, ADI-5~\cite{adi5} and ADI-17 \cite{shonADI17FineGrainedArabic2020, mgb5} shared tasks have been organized as part of the MGB challenge. ADI-5 contains 74 hours of audio segments from the Al Jazeera TV channel, classified into five dialect groups.
ADI-17 consists of 3,033 hours of audio segments from YouTube programs covering a variety of different genres. The segments are split into 17 different dialectal categories, allowing for much finer grain dialectal analysis than the ADI-5 corpus.
These efforts provided the research community with two valuable datasets for speech dialect identification, each offering different levels of granularity.

Initial approaches for ADI relied on acoustic features and classical machine learning methods like Support Vector Machines~\cite{wray2018classification} and Deep Neural Networks~\cite{lulu2018automatic}.
More recently, Self-Supervised Learning (SSL) has emerged as a promising approach in the domain of speech processing. SSL models like Hubert~\cite{hsu2021hubert} and wav2vec 2.0~\cite{baevski2020wav2vec}, which leverage large amounts of unlabeled data to learn representations without the need for explicit annotations, have demonstrated considerable potential in improving performance for many speech processing tasks for Arabic \cite{djanibekov2024dialectal, mdhaffar-etal-2024-taric, mdhaffar2024performance}, including ADI~\cite{waheed2023voxarabica}. 

In this work, we tackle the problem of the robustness of country-level Arabic speech dialect identification, making the following contributions: (1) we expand ADI-17's dialectal coverage by introducing the ADI-20 dataset, (2) we investigate the impact of training data quantity and model complexity on ADI model performance and (3) we release high-performing ADI models for the research community\footnote{Pre-trained models, dataset manifests, and complete recipes for ADI-17 and ADI-20 systems are available at \url{github.com/elyadata/ADI-20}}. According to our knowledge, this is the first publicly available ADI model.

\section{Datasets} \label{sec:dataset}

In this work, we use two datasets, namely ADI-17 and ADI-20, both of which are detailed in the following subsections.

\subsection{ADI-17} \label{subsec:dataset-adi17}

The official ADI-17 dataset~\cite{shonADI17FineGrainedArabic2020} contains 3,033 hours of dialectal Arabic speech for training and around 2 hours per dialect for the test and validation splits. 
However, the training data is imbalanced, with substantial differences between some dialects. For example, the quantity of training data for the Iraqi dialect is 31 times higher than the quantity available for Jordanian.

\begin{figure}[!htb]
    \centering
    \resizebox{\columnwidth}{!}{ 
        \begin{tikzpicture}
    \begin{axis}[
        width=\columnwidth, % Keeps the width consistent
        height=0.225\textheight, % Reduce the height (adjust the value as needed)\textbf{}
        ybar stacked,
        symbolic x coords={ALG, BAH, EGY, IRA, JOR, KSA, KUW, LEB, LIB, MAU, MOR, MSA, OMA, PAL, QAT, SUD, SYR, TUN, UAE, YEM},
        xtick=data,
        x tick label style={font=\scriptsize, rotate=90, anchor=east},
        y tick label style={font=\scriptsize},
        enlarge x limits=0.05,
        bar width=8pt,
        ymin=0, ymax=850,
        legend pos=north east,
        legend style={font=\scriptsize},
        every node near coord/.append style={font=\scriptsize},
        extra y ticks={53},
        extra y tick labels={53},
        extra y tick style={tick label style={font=\scriptsize, color=purple}}
    ]
        % ADI-17 data
        \addplot[pattern=north east lines, pattern color=black, area legend] coordinates {
            (ALG,115.78005)
            (BAH,0)
            (EGY,451.1721056)
            (IRA,815.8002611)
            (JOR,25.97206667)
            (KSA,186.1909222)
            (KUW,108.2218722)
            (LEB,116.8250833)
            (LIB,127.4901389)
            (MAU,456.4925556)
            (MOR,57.89163333)
            (MSA,0)
            (OMA,58.58756111)
            (PAL,121.4282667)
            (QAT,62.31474444)
            (SUD,47.78931667)
            (SYR,119.5772611)
            (TUN,0)
            (UAE,108.4672111)
            (YEM,53.45601667)
        };
        \addlegendentry{ADI-17}

        % Added for ADI-17-53 and ADI-20
        \addplot[pattern=dots, pattern color=blue, area legend] coordinates {
            (ALG,0) (BAH,270.89) (EGY,0) (IRA,0) (JOR,27.52) (KSA,0) (KUW,0) (LEB,0) (LIB,0) (MAU,0)
            (MOR,0) (MSA,68.38) (OMA,0) (PAL,0) (QAT,0) (SUD,13.34) (SYR,0) (TUN,142.69) (UAE,0) (YEM,0)
        };
        \addlegendentry{Added for ADI-20}
        
        \draw [purple, thick, dashed] (rel axis cs:0,0.06235) -- (rel axis cs:1,0.06235);
  
    \end{axis}
\end{tikzpicture}
    }
    \caption{Distribution of ADI-17 and ADI-20 train data (hours) per country dialect. The horizontal line indicates the 53-hour threshold for ADI-17-53h and ADI-20-53h.}
    \label{fig:data-durations}
\end{figure}
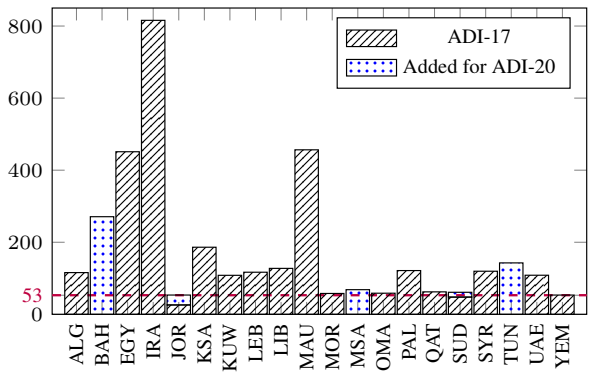

As shown in Figure~\ref{fig:data-durations}, Iraqi, Egyptian, and Mauritanian (Hassaniya) dialects combined account for over half the full ADI-17 training set. Therefore, some dialects, such as Jordanian and Sudanese dialects, are poorly represented in the training set. Even more importantly, we noticed that MSA and some Arabic dialects were not considered during the ADI-17 data collection process, namely Tunisian and Bahraini. For all these reasons, we proposed ADI-20, described in section \ref{adi20}.

\subsection{ADI-20}
\label{adi20}
The ADI-20 data set is proposed to broaden the dialectal coverage of ADI-17 and include MSA. First, we incorporated MSA to differentiate between regional dialects and the more formal, standardized Arabic. Second, Tunisian and Bahraini dialects were added to the dataset. Obtaining audio data for dialects from countries like Somalia and Eritrea remains challenging and is left for future exploration. Figure~\ref{fig:data-durations} illustrates the differences in dialectal coverage between ADI-17 and ADI-20.
For MSA, we used audio files from ADI-5 coupled with some scraped broadcast news data from YouTube. Overall, ADI-20 contains 68 hours of MSA speech. For evaluation sets, we have decided to use the same MSA dev and test sets used in ADI-5.
Regarding the Tunisian dialect, the whole TunSwitch~\cite{tunswitch} dataset was used, including the code-switched subset, for a combined duration of 142.7 hours. The train and dev splits were also based on TunSwitch splits but were augmented with content scraped from YouTube to reach 2 hours for each, as is the case for the other dialects. Lastly, Bahraini content is composed exclusively of content sourced from YouTube, with approximately 271 hours for training and 2 hours for validation and testing each.

In addition to the entirely new dialects, we collected more data for the two most underrepresented ADI-17 dialects: Sudanese and Jordanian. Both have been increased to reach the 53-hour training data of the Yemeni dialect.
In total, ADI-20 covers 20 Arabic dialects with at least 53 hours of training data per dialect and a total duration of \textbf{3556.76 hours} of MSA and dialectal Arabic speech. 

\subsection{Data preparation} \label{subsec:dataset-data-preparation}

Since this work aims to assess the impact of training data quantity on ADI model performance, we created three size-based variations of the original ADI-17 dataset described below: 
\begin{enumerate}
    \item \textbf{ADI-17-10h}: Limiting training data to 10 hours per dialect. In this configuration, we are using only 5.6\% of the full ADI-17 train set.
    \item \textbf{ADI-17-25h}: 25 hours per dialect. In this setup, only 14\% of the full ADI-17 train set will be used for model training.
    \item \textbf{ADI-17-53h}: 53 hours per dialect. This is equivalent to the usage of about 30\% of the total duration of the full training set. Note that we still use only 17 dialects except for Jordanian and Sudanese dialects, which we completed with data from ADI-20 to reach the target size of 53 hours per dialect.
    \item \textbf{ADI-17-full}: refers to the setup in which the entire ADI-17 dataset is used as it is and without any additional data.
\end{enumerate}

As a result of this data preparation step, we are able to explore different ADI models' performance with respect to the quantity of training data. It is worth mentioning that all three subsets are sampled using the same seed. We would also like to emphasize that the stratification considers the speech duration per dialect irrespective of the number of segments, as those can still differ between dialects. Moreover, all segments exceeding 30 seconds are split into smaller segments of 30 seconds or shorter, and those shorter than 3 seconds are discarded for all train datasets. For the scraped content, we rely on YouTube captions timestamps for audio segmentation.

On top of the 4 subsets described above, we created two additional subsets that integrated the newly introduced dialects into ADI-20. We refer to the first subset as \textbf{ADI-20-53h}: Created as an extension of \textbf{ADI-17-53h} with an additional 53 hours of speech segments from MSA, Bahraini, and Tunisian dialects. \textbf{ADI-20-full} refers simply to the full ADI-20 data set.

\section{Models}

In this section, we discuss the various trained ADI models. We experimented with two model architectures: (1) ECAPA-TDNN and (2) Whisper-based identification systems. All our models are trained using the open-source SpeechBrain~\cite{speechbrainV1} toolkit. 

\subsection{ECAPA-TDNN}
Building on the Time-Delay Neural Network (TDNN) framework, ECAPA-TDNN (Emphasized Channel Attention, Propagation, and Aggregation in TDNN)~\cite{desplanquesECAPATDNNEmphasizedChannel2020} introduces several enhancements, such as residual connections, squeeze-and-excitation modules, and channel-dependent attentive statistical pooling. The model also leverages Res2Blocks (residual blocks with dilated convolutions) to capture multiscale speech features while integrating position-aware attention to refine feature representations. These improvements enable ECAPA-TDNN to extract highly discriminative speech features, making it a strong candidate for classification tasks like dialect identification. ECAPA-TDNN has achieved state-of-the-art (SoTA) results on the ADI-17 dataset~\cite{kulkarniAnotherModelArabic2023}, which motivates its choice as a baseline system in this study. All our ECAPA-TDNN systems are implemented following the CommonLanguage recipe\footnote{\label{fn:SB-recipe} github.com/speechbrain/speechbrain/tree/develop/\\recipes/CommonLanguage}.
We also used the pre-trained model available on HuggingFace\footnote{{huggingface.co/speechbrain/spkrec-ecapa-voxceleb}} since it has been shown to give better results\footref{fn:SB-recipe}. A key difference from the original implementation is that 80-dimensional Mel filterbanks are used instead of MFCC features.

\subsection{Whisper encoder}
Whisper~\cite{whisper} is a weakly supervised family of encoder-decoder transformer models trained on 680,000 hours of audio crawled from the Internet, using a multitask objective that includes ASR, speech translation, and language identification. These models achieve near SoTA performance across multiple benchmarks and cover 96 languages beyond English, including Arabic.
Notably, studies such as~\cite{talafhaNShotBenchmarkingWhisper2023} have demonstrated Whisper's ability to deliver competitive results in ASR for MSA and particular Arabic dialects, even outperforming fine-tuned XLS-R models~\cite{babuXLSRSelfsupervisedCrosslingual2022} in some cases. Moreover, Whisper is available in multiple size variants (tiny, base, small, medium, large), which makes it a good choice to study model size scalability. Since ADI is fundamentally a classification task, we kept only the Whisper encoder block, and we replaced the decoder with an attention pooling mechanism followed by a classification layer.

\section{Experimental setup}

We consider three experimental configurations to address our research questions. All our models were trained on 4 GPUs with a batch size of 16.

\subsection{ECAPA-TDNN vs Whisper}
This setup is designed to compare ECAPA-TDNN and Whisper-based model performances for the ADI task. In order to ensure a fair comparison, we selected the Whisper-base model alongside ECAPA-TDNN. Both models are similar in size: 20.6 and 20.8 million parameters for Whisper-base and ECAPA-TDNN, respectively. Both models were trained using the same \textbf{ADI-17-full} without data augmentation.

\subsection{Data quantity vs model complexity}
The main purpose of this setup is to assess the impact of training data quantity and model size on model performance. For this purpose, we used three pre-trained Whisper models: Whisper small with 88.2M parameters, Whisper medium with 307.2M parameters, and Whisper large with 637 million parameters. All these models are trained on the \textbf{ADI-17-full} dataset and its three subsets, described in section~\ref{subsec:dataset-data-preparation}. 

\subsection{ADI-20 systems}
\label{adi20-sys}

Based on the results from the previously described setup, the third one aims to find the best possible model configuration to obtain the best ADI system. Several ECAPA-TDNN and Whisper models (small, medium, and large) were trained and evaluated on ADI-20 evaluation sets. Therefore, we used data augmentation methods and experimented with different layer-freezing approaches. 

\begin{table*}[htb]
\centering
\renewcommand\thetable{2}
    \caption{Weighted F1-scores of ADI-17 and ADI-20 models on their respective test sets. ADI-20 models are also evaluated on the ADI-17 test split. Bold indicates our best overall system.}
    \begin{tabular}{@{}l|cccc|cccc@{}}
        \toprule
        \multicolumn{1}{c|}{Train dataset} &
          \begin{tabular}[c]{@{}c@{}}ADI-17\\ Full\end{tabular} &
          \begin{tabular}[c]{@{}c@{}}ADI-17\\ 53h\end{tabular} &
          \begin{tabular}[c]{@{}c@{}}ADI-17\\ 25h\end{tabular} &
          \begin{tabular}[c]{@{}c@{}}ADI-17\\ 10h\end{tabular} &
          \multicolumn{2}{c}{\begin{tabular}[c]{@{}c@{}}ADI-20\\ 53h\end{tabular}} &
          \multicolumn{2}{c}{\begin{tabular}[c]{@{}c@{}}ADI-20-53h\\ + frz. + aug.\end{tabular}} \\ \midrule
        \multicolumn{1}{c|}{Test dataset} & \multicolumn{4}{c|}{ADI-17}                             & ADI-17 & ADI-20 & ADI-17 & ADI-20 \\ \midrule
        HuBERT-17 ~\cite{casablanca} & 92.12 & \_ & \_  & \_ &\_ & \_ & \_& \_   \\ \midrule
        Whisper-small                     & 93.80                       & 94.42 & 92.86 & 89.52 & 93.52  & 90.88  & 93.45  & 91.54  \\
        Whisper-med.                      & 95.46               & 95.29 & 94.10  & 92.88  & 94.96  & 93.59  & 95.03  & 93.39  \\
        Whisper-large                     & 95.66    & \underline{95.16}       & 93.91 & 92.96 & \underline{95.41}  & 94.79  & \textbf{95.82}  & \textbf{94.83}  \\ \bottomrule
    \end{tabular}
    \label{tab:adi17-adi20-test}
\end{table*}

\section{Results} \label{sec:results}

Table~\ref{tab:adi17-adi20-test} presents the weighted F1-scores achieved by our ADI systems. The first half of the table reports results for models trained on ADI-17 subsets and evaluated on ADI-17 test sets. The second half shows results for ADI-20 systems evaluated on both ADI-17 and ADI-20 test splits. Table~\ref{tab:ecapa-whisper-base-results} reports evaluation results of ECAPA-TDNN systems and the Whisper-base system.

\subsection{ECAPA-TDNN vs Whisper}
Table~\ref{tab:ecapa-whisper-base-results} presents the dialect identification results on ADI-17 test sets using both ECAPA-TDNN and the Whisper-base models. Obtained results demonstrate that ECAPA-TDNN outperforms the Whisper-based system when they have a similar number of parameters: A weighted F1-score of 93.16\% is obtained with ECAPA-TDNN, while the Whisper-based model obtains only 91.34\%. We believe this could be explained by the fact that ECAPA-TDNN, unlike the Whisper-base model, was pre-trained on a speaker recognition task, which is somehow similar to dialect identification.

\begin{table}[H]
\renewcommand\thetable{1}
    \centering
    \setlength{\tabcolsep}{3pt}
    \renewcommand{\arraystretch}{1.0}
    \caption{Weighted F1-scores of ADI using Whisper-base and ECAPA-TDNN models trained using \textbf{ADI-17-full} and testing on ADI-17 test sets.}
    \begin{tabular}{@{}l|c@{}}
        \toprule
        \multicolumn{1}{c|}{} &  Weighted F1-score \\ 
        \midrule
        ECAPA-TDNN & 93.16  \\ 
        Whisper-base & 91.34 \\
        \bottomrule
    \end{tabular}
    \label{tab:ecapa-whisper-base-results}
\end{table}

\subsection{Data quantity vs model complexity}
We carried out several experiments in order to evaluate the impact of training dataset quantity and model size on dialect identification model performance. As previously mentioned, Whisper-based systems were chosen for their multiple-size option, which is aligned with our goals to investigate model scaling.

First, we experimented with using a bigger Whisper model to replace the Whisper-base used in the previous section. As shown in the results presented in the first column of Table~\ref{tab:adi17-adi20-test}, the F1-score improves with increasing the Whisper model size. A better F1-score of 95.66\% is achieved compared to the 91.34\% obtained with Whisper-base and reported in Table~\ref{tab:ecapa-whisper-base-results}. This shows that, as expected, larger models outperform smaller ones for our ADI task. ~\\

Next, we focused on training different ADI models with smaller amounts of training data. The results of this experiment are summarized in Table~\ref{tab:adi17-adi20-test}. As we can see from the table, using only 10 hours of training data per dialect, only 5.6\% of the total ADI-17 training data, significantly decreases the ADI system performance irrespective of the model size. For instance, for the Whisper-large model, we obtained an F1-score of 92.96\% with ADI-17-10h while ADI-17-Full gives  95.66\%. In the same vein, Table~\ref{tab:adi17-adi20-test} gives the F1-score of models trained using ADI-17-25h and ADI-17-53h. As shown in the table, the gap gets narrower, and we were even able to get a better F1-score for the Whisper-small model using ADI-17-53h of 94.42\%, compared to 93.80\% when using ADI-17-Full (i.e., with only 30\% of the original training data).

Finally, it is worth mentioning that Whisper-medium achieves a similar F1-score compared to Whisper-large (95.29\% vs 95.66\%) model despite having half as many parameters and only using 30\% of the training data.

\subsection{ADI-20 systems results}

Based on these previous observations, we moved forward to the ADI-20 setup presented in section \ref{adi20-sys}. In this context, we trained various ADI systems using the ADI-20-53h training subset. The results are presented in the last column of Table~\ref{tab:adi17-adi20-test}. Note that the obtained models are evaluated on both ADI-17 and ADI-20 test sets. This is done for comparability reasons with models trained and evaluated using only ADI-17 (i.e. column 2 in Table~\ref{tab:adi17-adi20-test}). 

As we can see from the reported results, we obtained slightly better results on the ADI-17 test set by training Whisper-large with ADI-20-53h rather than ADI-17-53h: 95.41\% of F1-score for the former versus 95.16\% of F1-score for the latter (see underlined scores in Table \ref{tab:adi17-adi20-test}). Note that this improvement only manifests with Whisper-large architecture: Whisper-small and Whisper-medium systems trained on ADI-20-53h don't outperform their counterparts trained on ADI-17-53h. We also carried out other sets of experiments, in order to push further the ADI models trained using ADI-20-53h data set. Mainly we experimented with data augmentation and encoder lower layer freezing since it has been shown to be helpful for other speech processing tasks {\cite{eberhard-zesch-2021-effects, zanon-boito-etal-2022-trac}. Overall, the best results are obtained using Whisper-large system by freezing the first half of its encoder layers and using data augmentation as implemented in the Speechbrain toolkit \cite{speechbrainV1}.\\

Finally, we also trained the ECAPA-TDNN model using the ADI-20-53h dataset, and the obtained model gives an F1-score of respectively 92.89\% and 90.49\% and ADI-17 and ADI-20 test sets.

\subsection{Zero-shot evaluation}

We further evaluate our best systems on the recently released Casablanca dataset~\cite{casablanca} in a zero-shot fashion. Results presented in Table~\ref{tab:eval-casablanca}, show that Whisper large trained with ADI-20-53h+frz+aug achieves an F1-score of 62.74\%, significantly outperforming the 39.24\% reported in~\cite{casablanca} for the system described in~\cite{sullivanRobustnessArabicSpeech2023}, which was also evaluated in a zero-shot manner.\\

\begin{table}[htb]
\centering
\renewcommand\thetable{3}
    \caption{Weighted F1-scores of ADI-17 and ADI-20 systems zero-shot evaluation on the Casablanca test set. Bold indicates the best overall system.}
    \begin{tabular}{@{}l|cc}
    \toprule
    \multicolumn{1}{c|}{\diagbox{Models}{Train \\ data}}&
      \begin{tabular}[c]{@{}c@{}}ADI-17\\ Full\end{tabular} &
      \begin{tabular}[c]{@{}c@{}}ADI-20\\53h\\ frz+aug\end{tabular} \\ \midrule
    HuBERT-17 \cite{casablanca}    &  39.24              & \_       \\ \midrule
        Whisper med. & 53.84        & 58.11          \\
        Whisper large  & 58.89      & \textbf{62.74} \\ \bottomrule
    \end{tabular}
    \label{tab:eval-casablanca}
\end{table}

According to the results reported in Table \ref{tab:eval-casablanca}, the best results are obtained with ADI-20-53h with layer freezing and data augmentation. We hypothesize that augmentations enhance robustness to the unseen conditions of TV dramas, which differ from the domains of our training datasets consisting of YouTube-sourced videos.\\

We also evaluated our ADI-17-full system, which is trained using the same training data as the HuBERT-17 model reported in \cite{casablanca}. Both Whisper-medium and Whisper-large models clearly outperform HuBERT-17 with an F1-score of 53.84\% and 58.89\%, respectively. An error analysis of the results reveals that most errors happen when classifying Jordanian, which is often mis-classified as Egyptian or Syrian. Moreover, a considerable amount of misclassification happens between Algerian, Moroccan, and Libyan (which are Maghrebi Arabic dialects).

\subsection{Error analysis}
Error analysis reveals frequent misclassifications between geographically or linguistically close dialects, e.g., Jordanian as Lebanese/Palestinian/Syrian and Bahraini as Emirati/Qatari. 
MSA, though not a dialect, is primarily confused with North African varieties (Algerian, Libyan, Egyptian), challenging expectations given Arabic’s origins in the Arabian Peninsula.

\section{Conclusion}
This paper presents ADI-20, an extension of ADI-17 designed for country-level Arabic dialect identification, offering comprehensive coverage of dialects from almost all Arabic-speaking countries.
Using this dataset and the ADI-17 dataset, we investigate the impact of training data quantity and model complexity on ADI model performance, and we demonstrate that competitive ADI models can be trained with as little as 53 hours of data per dialect.
We release our data, models, and training recipes to the community to facilitate further research and reproducibility.  
For future work, we plan to investigate ECAPA-TDNN models further, which demonstrated promising results but were not explored in depth due to time constraints and the availability of Whisper models in various sizes. Additionally, we aim to extend our research to city-level Arabic dialect identification, capturing finer linguistic variations within countries to improve dialect classification granularity.

\section{Acknowledgements}
This work was partially funded by the ESPERANTO project. The ESPERANTO project has received funding from the European Union’s Horizon 2020 (H2020) research and innovation program under the Marie Skłodowska-Curie grant agreement No 101007666.
This work was granted access to the HPC resources of IDRIS under the allocations AD011015051, AD011012551R3, and AD011012108R made by GENCI. \\

\bibliographystyle{IEEEtran}
\bibliography{mybib}

\end{document}